\pdfoutput=1

\documentclass[11pt]{article}

\usepackage[]{EMNLP2022}

\usepackage{times}
\usepackage{latexsym}

\usepackage[T1]{fontenc}

\usepackage[utf8]{inputenc}

\usepackage{array}
\usepackage{caption}
\usepackage{booktabs}
\usepackage{subcaption}
\usepackage{graphicx}
\usepackage{multirow}

\usepackage{microtype}

\usepackage{inconsolata}

\title{An Empirical Revisiting of Linguistic Knowledge Fusion in \\ Language Understanding Tasks}

\author{Changlong Yu\textsuperscript{$1$} \quad Tianyi Xiao\textsuperscript{$1$} \quad  Lingpeng Kong\textsuperscript{$2$} \quad Yangqiu Song\textsuperscript{$1$} \quad Wilfred Ng\textsuperscript{$1$} \\
\textsuperscript{$1$}HKUST, Hong Kong \quad \textsuperscript{$2$}The University of Hong Kong, Hong Kong \quad \\
 \{cyuaq, yqsong, wilfred\}@cse.ust.hk, \ txiao@connect.ust.hk, \ lpk@cs.hku.hk\\ 
}

\begin{document}
\maketitle
\begin{abstract}

Though linguistic knowledge emerges during large-scale language model pretraining, recent work attempt to explicitly incorporate human-defined linguistic priors into task-specific finetuning. Infusing language models with syntactic or semantic knowledge from parsers has shown improvements on many language understanding tasks. 
To further investigate the effectiveness of structural linguistic priors, we conduct empirical study of replacing parsed graphs or trees with trivial ones~(rarely carrying linguistic knowledge e.g., balanced tree) for tasks in the GLUE benchmark.
Encoding with trivial graphs achieves competitive or even better performance in fully-supervised and few-shot settings.
It reveals that the gains might not be significantly attributed to explicit linguistic priors but rather to more feature interactions brought by fusion layers.
Hence we call for attention to using trivial graphs as necessary baselines to design advanced knowledge fusion methods in the future.  

\end{abstract}

\section{Introduction}

Recently large-scale pretrained language models~\cite{devlin-etal-2019-bert,liu2019roberta,2020t5} have shown to gain linguistic knowledge from unlabeled corpus and achieve strong performance on many downstream natural language processing~(NLP) tasks.
Though probing analysis indicate that, to some extent, they can implicitly capture syntactic or semantic structures~\cite{hewitt-manning-2019-structural,Goldberg2019AssessingBS,tenney2018you,hou-sachan-2021-birds}, whether they can further benefit from more explicit linguistic knowledge remains an open problem. 
Attempts have been made to inject syntactic biases into language model pretraining~\citep{Kuncoro2020SyntacticSD,wang-etal-2021-k,xu-etal-2021-syntax} or infuse finetuning with semantic information~\cite{zhang2020semantics,wu2021infusing}, and positive results are reported on downstream tasks.

However, the concerns about the effect or viability of linguistic knowledge have been raised. On the one hand, the performance gains highly rely on the availability of human-annotated dependency parsers~\cite{sachan-etal-2021-syntax} or oracle semantic graphs~\cite{Prange2021LinguisticFG}, which limits the real-world applications. Developing accurate semantic graph parsers is yet challenging~\cite{oepen-etal-2019-mrp,bai-etal-2022-graph}. 
On the other hand, incorporating trees induced from pretrained language models~\cite{wu-etal-2020-perturbed} can outperform the ones fused with dependency-parsed trees for aspect-level sentiment analysis~\cite{dai-etal-2021-syntax}. 
This  discovery is in line with the similar findings of trivial trees for tree-LSTM encoders in sequence modeling tasks~\cite{shi2018tree}. 
In this work, we push the envelop and answer the following two questions. Do knowledge fusion methods in \citet{wu2021infusing} benefit from trivial graphs that contain no linguistic information? If that's the case, where might the performance gains come from?

With the above questions, we empirically revisit the effectiveness of linguistic knowledge fusion in language understanding tasks. Motivated by \citet{shi2018tree}, we compare the performance between original dependency-parsed trees and balanced trees for syntax fusion, and compare the results between parsed semantic graphs and sequential graphs for semantic fusion. 
To our surprise, trivial graphs outperform syntactic trees or semantic graphs in full-supervised setting and achieve competitive results in few-shot setting. 
All the evidence again shows that the linguistic inductive bias might not be the major contributor of consistent improvements over baselines. 
Additional analysis gives some clues that the possible reasons are extra model parameters and feature interactions from fusion modules.
This work encourages future research to add trivial graphs as necessary baselines when designing more advanced knowledge fusion methods for downstream tasks. 
Our experimental code is available at \url{https://github.com/HKUST-KnowComp/revisit-nlu-linguistic-knowledge}.


\section{Study Design} 

In this section, we briefly introduce two linguistic graphs, i.e., syntactic dependency trees and semantic graphs. As a comparison, we manually construct two trivial graphs to infuse with task-specific finetuning.

\subsection{Linguistic Graph}

Graphs have intuitively represented various linguistic phenomena in natural languages including sentence structures~\cite{chomsky1957logical} and meanings~\cite{koller-etal-2019-graph}.  

\noindent \textbf{Syntactic Dependency Tree.} Syntactic trees are one of the most commonly-used linguistic structures and have long been shown useful for many NLP tasks. 
Syntactic dependency mainly models \textit{head-dependent} relations between words. Dependency parsers parse the sentence into tree structures, which are further incorporated into LMs via syntax-aware attention~\cite{nguyen2019tree} or graph neural networks~(GNN~\citealt{sachan-etal-2021-syntax}).

\begin{figure}
    \centering
    
    \includegraphics[scale=0.40]{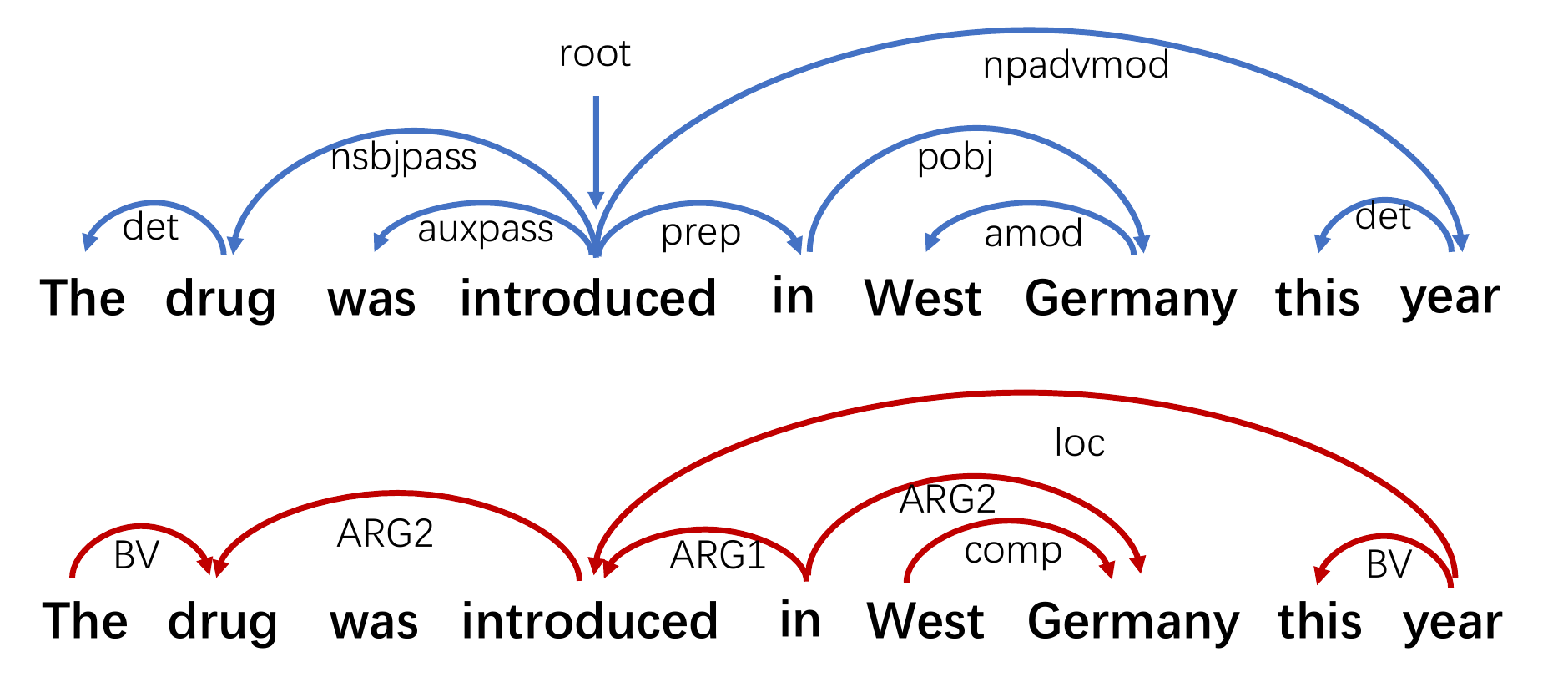}
    \caption{An example of dependency tree~(blue) and DM semantic graph~(red).}
    \label{fig:graph_example}
\end{figure}

\noindent \textbf{Semantic Graphs.} Different from syntactic dependency, semantic graphs aim to map sentences to high-order meaning representations with more complex structures. Normally semantics concern about \textit{predicate-argument} relations, where predicates evoke relations of various arity and arguments filled with semantic roles that are related to each specific predicate.\footnote{We refer the readers to the ACL tutorial \citet{koller-etal-2019-graph} for detailed explanations.} 
One example is shown in Figure~\ref{fig:graph_example}, and the characteristics of semantic graphs are the following: 1)~Argument sharing leads to nodes whose in-degrees are more than one. 2)~Some tokens do not contribute to meaning and not appear in the graphs. 3)~There exist multiple roots. 
Complex semantic structures enable them to capture information that is not explicit in the single-rooted syntactic trees. 
Semantics could be formalized by different frameworks with respect to special linguistic assumptions. Some representative semantic formalisms are AMR~(Abstract Meaning Representation,~\citealp{banarescu-etal-2013-abstract}) and UCCA~(\citealp{abend-rappoport-2013-universal}). 
Recently \citet{wu2021infusing} proposed semantics-infused finetuning~(SIFT) to infuse DM~(DELPH-IN Minimal Recursion Semantics,~\citealp{ivanova-etal-2012-contrastive}) graphs and achieved consistent improvements over RoBERTa~\cite{liu2019roberta} baselines on the GLUE~\cite{wang2019glue}.

DM graphs~\cite{ivanova-etal-2012-contrastive} define 59 types to characterize \textit{predicate-argument} relationships.
In order to investigate the effect of different semantic relations, we consider to only keep six common relation types, which appear in most parsed graphs, named \textbf{skeleton graphs}. 
These relations include \texttt{ARG1}, \texttt{ARG2}, \texttt{ARG3}, \texttt{ARG4}, \texttt{compound} and \texttt{BV}.  
We are interested in whether downstream tasks would still benefit from the core semantics instead of  the entire linguistic graphs.

\subsection{Trivial Graph}

Though linguistic graphs convey useful structures, high-quality parsers are not easily available due to limited annotated graph banks~\cite{oepen-etal-2019-mrp}. 
If structure priors are unavailable, \citet{shi2018tree} demonstrated that trivial trees, such as \textit{gumbel tree} outperform syntactic trees when they are incorporated into tree LSTM encoders~\cite{tai-etal-2015-improved}.

However, infusing trivial linguistic graphs with pretrained transformer models has not been explored. Similarly, we also create two types of trivial trees or graphs, which rarely contain linguistic inductive bias, to reproduce knowledge fusion experiments in \citet{wu2021infusing}.

\noindent \textbf{Binary Balanced Tree.} Compared with syntactic trees, binary balanced trees are shallower and possibly easier to propagate information from leaves to the root. We assume GNN layers might benefit from the shallowness of balanced trees.

\noindent \textbf{Sequential Bidirectional Graph.}
As the most natural and straight-forward way, tokens in the sentence are connected in the sequential order, which combines left-to-right and right-to-left chains. 
By doing so, GNN layers only aggregate local information rather than potentially long dependency from linguistic graphs. 

\begin{table*}[h]
\centering
\setlength\tabcolsep{4.5pt}
\begin{tabular}{lccccccccc}
\hline
\textbf{Models} & \textbf{CoLA} & \textbf{MRPC} & \textbf{STS-B} & \textbf{SST-2} & \textbf{RTE} & \textbf{QNLI} & \textbf{QQP} & \textbf{MNLI} & \textbf{Avg.} \\
\hline
\textsc{RoBERTa} & 63.1{\scriptsize $\pm$0.9} & 90.1{\scriptsize $\pm$0.8} & 91.0{\scriptsize $\pm$0.0} & 94.6{\scriptsize $\pm$0.3} & 79.0{\scriptsize $\pm$1.6} & 93.0{\scriptsize $\pm$0.3} & 91.8{\scriptsize $\pm$0.1} & 87.7{\scriptsize $\pm$0.2} & 86.3\\
\hline
\textsc{SIFT} & 64.8{\scriptsize $\pm$0.4} & 90.5{\scriptsize $\pm$0.7} & 91.3{\scriptsize $\pm$0.1} & 95.1{\scriptsize $\pm$0.4} & 81.0{\scriptsize $\pm$1.4} & 93.2{\scriptsize $\pm$ 0.2} & 91.9{\scriptsize $\pm$ 0.1} & 87.9{\scriptsize $\pm$ 0.2} & 87.0\\
\textsc{SIFT~(Ske)} & \textbf{65.6}{\scriptsize $\pm$0.8}  & 90.6{\scriptsize $\pm$0.8} & 91.4{\scriptsize $\pm$0.2} & 95.2{\scriptsize $\pm$0.2} & \textbf{81.8}{\scriptsize $\pm$0.3} &  93.2{\scriptsize $\pm$0.2} & 91.9{\scriptsize $\pm$0.0 } & 87.9{\scriptsize $\pm$0.0}  & 87.2 \\
$\textsc{SIFT~(Seq)}^{\spadesuit}$ & 65.5{\scriptsize $\pm$0.3}  & 90.9{\scriptsize $\pm$0.3} & 91.4{\scriptsize $\pm$0.0} & 95.2{\scriptsize $\pm$0.2} & 81.3{\scriptsize $\pm$1.8} & 93.2{\scriptsize $\pm$0.1} & 91.9{\scriptsize $\pm$0.1} & 87.9{\scriptsize $\pm$0.0 } & 87.2  \\
\hline
\textsc{Syntax} & 63.5{\scriptsize $\pm$0.6} & 90.4{\scriptsize $\pm$0.5} & 91.1{\scriptsize $\pm$0.2}  & 94.7{\scriptsize $\pm$0.5} & 80.9{\scriptsize $\pm$1.0} & 92.8{\scriptsize $\pm$0.2} & 91.8{\scriptsize $\pm$0.0} & 87.9{\scriptsize $\pm$0.1} & 86.6\\
$\textsc{Balanced}^{\heartsuit}$ & 65.3{\scriptsize $\pm$0.9} & 90.5{\scriptsize $\pm$0.2} & \textbf{91.5}{\scriptsize $\pm$ 0.1} & \textbf{95.4}{\scriptsize $\pm$ 0.3}  & 81.1{\scriptsize $\pm$0.7} & 93.2{\scriptsize $\pm$ 0.1} & 91.9{\scriptsize $\pm$ 0.1} & 87.9{\scriptsize $\pm$ 0.1} & 87.1\\
\hline
\multicolumn{10}{c}{ (a) Base.} \\ 
\multicolumn{10}{c}{} \\ 
\hline
\textbf{Models} & \textbf{CoLA} & \textbf{MRPC} & \textbf{STS-B} & \textbf{SST-2} & \textbf{RTE} & \textbf{QNLI} & \textbf{QQP} & \textbf{MNLI} & \textbf{Avg.} \\
\hline
\textsc{RoBERTa} & 68.0{\scriptsize $\pm$0.6} & 90.1{\scriptsize $\pm$0.8} & 92.3{\scriptsize $\pm$0.2} & 96.1{\scriptsize $\pm$0.3} & 85.1{\scriptsize $\pm$1.0} & 94.5{\scriptsize $\pm$0.2} & 91.9{\scriptsize $\pm$0.1} & 90.3{\scriptsize $\pm$0.1} & 88.5\\
\hline
\textsc{SIFT} & 69.7{\scriptsize $\pm$0.5} & 91.3{\scriptsize $\pm$0.4} & 92.6{\scriptsize $\pm$0.0} & 96.3{\scriptsize $\pm$0.3} & 87.0{\scriptsize $\pm$1.1} & 94.7{\scriptsize $\pm$ 0.1} & 92.1{\scriptsize $\pm$ 0.1} & 90.4{\scriptsize $\pm$ 0.1} & 89.3\\
\textsc{SIFT~(Ske)} & \textbf{71.6}{\scriptsize $\pm$0.3}  & 91.8{\scriptsize $\pm$0.3} & \textbf{92.6}{\scriptsize $\pm$0.1} & 96.4{\scriptsize $\pm$0.2} & 88.5{\scriptsize $\pm$1.2} &  \textbf{94.8}{\scriptsize $\pm$0.1} & 92.1{\scriptsize $\pm$0.1 } & 90.4{\scriptsize $\pm$0.2} & 89.8   \\
$\textsc{SIFT~(Seq)}^{\spadesuit}$ & 70.7{\scriptsize $\pm$0.7}  & \textbf{91.9}{\scriptsize $\pm$0.5} & \textbf{92.6}{\scriptsize $\pm$0.1} & 96.5{\scriptsize $\pm$0.3} & \textbf{88.7}{\scriptsize $\pm$0.3} & 94.7{\scriptsize $\pm$0.1} & 92.1{\scriptsize $\pm$0.1} & 90.4{\scriptsize $\pm$0.1}  & 89.7 \\ 
\hline
\textsc{Syntax} & 69.6{\scriptsize $\pm$1.2} & 91.0{\scriptsize $\pm$0.5} & 92.4{\scriptsize $\pm$0.1}  & 95.9{\scriptsize $\pm$0.3} & 86.0{\scriptsize $\pm$1.6} & 94.6{\scriptsize $\pm$0.1} & 92.0{\scriptsize $\pm$0.0} & 90.4{\scriptsize $\pm$0.3} & 89.0\\

$\textsc{Balanced}^{\heartsuit}$ & 70.5{\scriptsize $\pm$1.0} & 91.5{\scriptsize $\pm$0.4} & 92.6{\scriptsize $\pm$ 0.0} & 96.5{\scriptsize $\pm$ 0.2}  & 88.0{\scriptsize $\pm$0.3} & 94.7{\scriptsize $\pm$ 0.1} & 92.1{\scriptsize $\pm$ 0.1} &  90.4{\scriptsize $\pm$0.1} & 89.5\\
\hline
\multicolumn{10}{c}{ (b) Large.} \\ 
\end{tabular}
\caption{\label{tab:main_results}
GLUE benchmark development results using RoBERTa base~(top) and large~(bottom). $\spadesuit$: we replace parsed semantic graphs~(\textsc{SIFT}) with trivial sequential graphs. $\heartsuit$: we replace parsed dependency trees~(\textsc{Syntax}) with balanced trees. \textsc{SIFT~(SKE)} means encoding with skeleton graphs. Trivial graphs could outperform parsed linguistic graphs in both base and large models.}
\end{table*}

\subsection{Encoding Graph Structures}
\label{sec:graph_encode}

Structural information can be incorporated into pretrained transformer models by two typical  strategies: adopt GNN on top of the output of transformers~\cite{wu2021infusing,peng-etal-2021-structure} and fuse structures with transformer attention layers~\cite{nguyen2019tree,zhang2019sgnet}.  
Followed \citet{wu2021infusing}, in this work we use the former one, where the linguistic effects are easily disentangled for analysis.  

Formally, given an input sentence $x_i=\{w_1, w_2, ..., w_L\}$ with the length $L$ and the corresponding graph $\mathcal{G}_i$~(either linguistic graph or trivial graph), we obtain the last hidden representation, $\mathbf{H}$ after pretrained transformer layers~\cite{vaswani2017attention}, which also serves as the node embedding initialization of relational graph neural networks~(RGCN~\citealt{schlichtkrull2018modeling}) to encode $\mathcal{G}_i$. At RGCN layers, each node's representation would be updated by aggregating its neighbors' features with relational bias. We max-pool over the final RGCN layer's output as the graph representation $\mathbf{O}_g$:
\begin{equation}
    \mathbf{O}_{g} = \mathrm{MaxPooling}(\mathrm{RGCN}([ \mathcal{G}_i, \mathbf{H}])).
\end{equation}
And the final classification feature is the concatenation of \texttt{[CLS]} token embedding $\mathbf{H}_0$ and pooled graph representation $\mathbf{O}_g$. Note that vanilla transformer-based models only take the \texttt{[CLS]} embedding as the classification feature. 
For sentence pair tasks, two graphs are separately encoded by RGCN with inner-attention and then aggregated to one representation as \citet{wu2021infusing}.  

\section{Experiments}

\subsection{Implementation Details}

We use the GLUE benchmark~\cite{wang2019glue}, a general natural language understanding test suite which contains eight datasets for text classification tasks~(Details listed in Appendix~\ref{sec:app_data_stat}). 
Following the common practice, we report the averaged results of development sets over multiple seeds.

We directly adopt parsed semantic graphs from  \citet{wu2021infusing}\footnote{\url{https://github.com/ZhaofengWu/SIFT}} for each dataset, using the ranked-first parser~\cite{che-etal-2019-hit} in the CoNLL2019 shared task~\cite{oepen-etal-2019-mrp}. 
For the construction of trivial graphs, we randomly sample edge labels from the relation list due to their unavailability. 
To be comparable with \citet{wu2021infusing}, we apply the same model architecture and tuned the same set of parameters such as learning rate, RGCN graph layer, RGCN hidden dimension, when infusing graph structures into LMs.

\subsection{Main results}

Table~\ref{tab:main_results} shows the main results of infusing with linguistic and trivial graphs. We can draw the following observations. 
First, sequential graphs and balanced trees achieve consistent improvements over corresponding linguistic ones. Also trivial graphs further improve RoBERTa baselines by more than 1.0 averaged point.  
Moreover, the random relation types in the trivial graphs have no effect on performance, again suggesting that linguistic structures and relations are not major contributors of the improvements.  
Second, skeleton graphs surprisingly improve the whole linguistic graphs, which indicates the fine-grained sentential semantics might not be necessary for language understanding tasks in GLUE.
Note that skeleton graphs keep 75\%-90\% edges of parsed graphs on average~(Detailed statistics in Appendix~\ref{sec:skeleton}).

\begin{figure}[t]
\centering
\subfloat[Few-shot experiments for the CoLA dataset]{\label{fig:cola}
\includegraphics[scale=0.43]{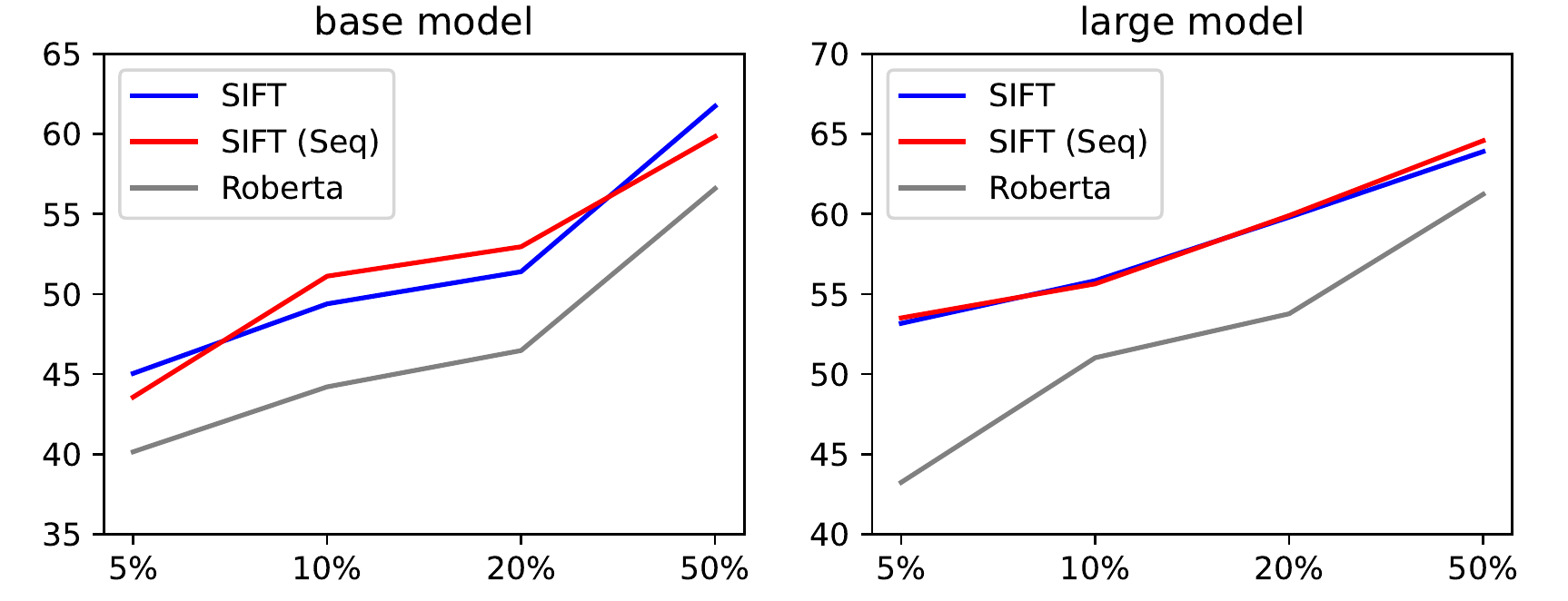}}
\
 \subfloat[Few-shot experiments for RTE dataset]{\label{fig:rte}
\includegraphics[scale=0.43]{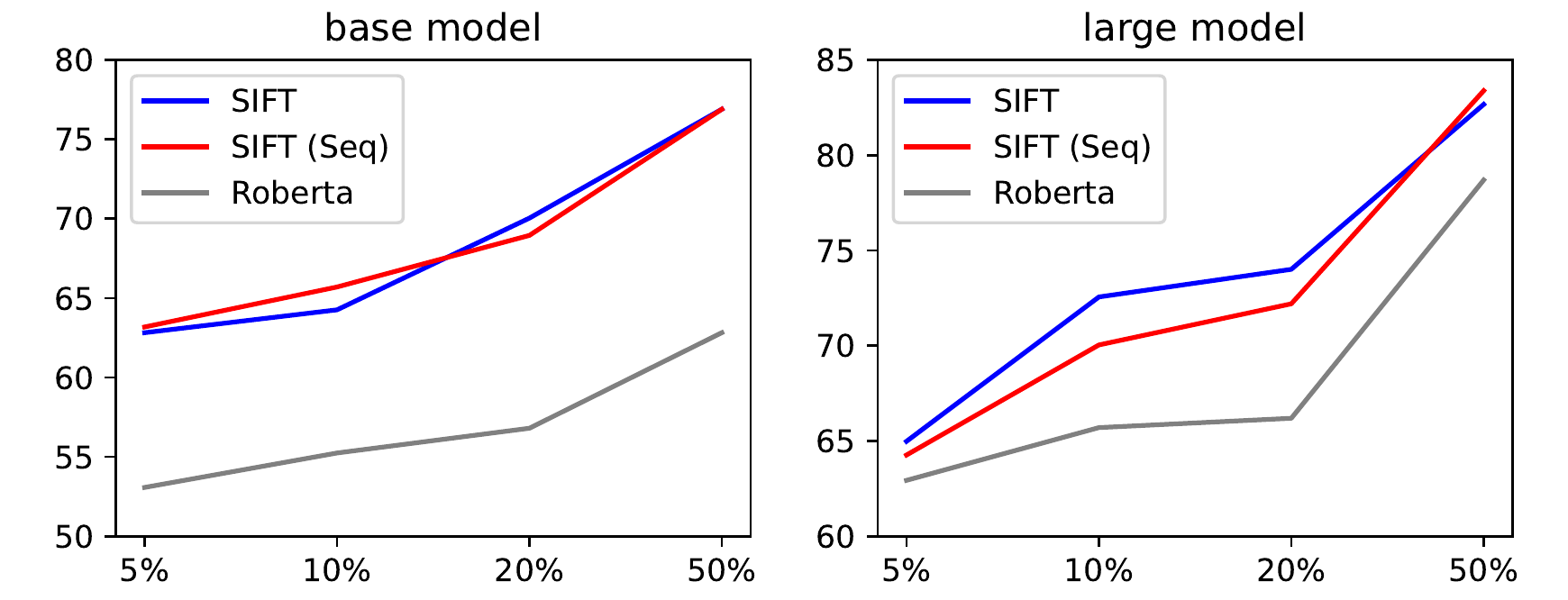}}
\caption{Performance comparison of RoBERTa, \textsc{SIFT} and \textsc{SIFT~(Seq)} with different training data sizes. The rest of datasets follow similar trends.}
\label{fig:compare}
\end{figure}

To examine the impact of trivial graphs with different sizes of training data, especially in low-resource scenarios, we randomly sample 5\%, 10\%, 20\% and 50\% training data. The trivial graphs yield competitive or even better results for CoLA and RTE datasets in Fig~\ref{fig:compare} across different sizes.

\subsection{Effect of Model Components}

Given the strong performance of trivial graphs, we conduct the ablation study of model components in Section~\ref{sec:graph_encode} besides graph structures. 
We first investigate whether pooling over transformer outputs can replace graph encoders, i.e., passing zero layer RGCN.
The results in Table~\ref{tab:ablation} show that the concatenation of \texttt{CLS} embedding and  direct pooled hidden states still falls behind graph fusion methods.
But one interesting finding is that such simple way by introducing token embedding features can have large improvements over RoBERTa baselines.

Second, RGCN layers over sequential graphs achieve good performance in Table~\ref{tab:main_results} . Considering the nature of sequential modeling ability of transformers, one question is whether additional transformer layers over sequences could learn better graph representations and improve the performance. 
The difference between transformer and RGCN is that the transformer captures complete graphs while RGCN takes sequential graphs as inputs.
We stack more randomly-initialized transformer layers over the pretrained encoder outputs~(comparable number of parameters with RGCN).  
Training with additional transformer layers yields similar results with RGCN encoders. 
From this perspective, structure biases make little difference and the gains of trivial graphs might be the results of additional token embedding features as well as their interactions via fusion modules~(RGCN or transformer).



\begin{table}[t]\small
\centering
\setlength\tabcolsep{4pt}
\begin{tabular}{l|l|c|c}
\toprule
 Dataset  & Method &  Base  &  Large      \\ \midrule 
\multirow{4}{*}{\begin{minipage}{0.2in}CoLA\end{minipage}}& \texttt{CLS}~(RoBERTa)  & 63.1{\scriptsize $\pm$0.9}    &  68.0{\scriptsize $\pm$0.6}     \\
&  \texttt{CLS} + RGCN~(SIFT)  & 64.8{\scriptsize $\pm$0.4} &  69.7{\scriptsize $\pm$0.5} \\ 
&  \texttt{CLS} + 0 RGCN & 64.5{\scriptsize $\pm$0.9}  &  69.4{\scriptsize $\pm$0.9}    \\ 
&  \texttt{CLS} + transformer & 64.8{\scriptsize $\pm$0.6}  &  70.1{\scriptsize $\pm$0.3}  \\
\midrule
\multirow{4}{*}{\begin{minipage}{0.2in}RTE\end{minipage}}& \texttt{CLS}~(RoBERTa)  & 79.0{\scriptsize $\pm$1.6}  &   85.1{\scriptsize $\pm$1.0}    \\
&  \texttt{CLS} + RGCN~(SIFT) &  81.0{\scriptsize $\pm$1.4} &  87.0{\scriptsize $\pm$1.1}     \\
&   \texttt{CLS} + 0 RGCN & 80.4{\scriptsize $\pm$0.9}&  86.7{\scriptsize $\pm$0.4}     \\
&   \texttt{CLS} + transformer &  81.2{\scriptsize $\pm$0.7}  &  87.1{\scriptsize $\pm$0.2} \\

\bottomrule
\end{tabular}
\caption{\label{tab:dataset} Results of different variants for \textsc{SIFT}. ``\texttt{CLS} + 0 RGCN'' means direct pooling without graph encoders while ``\texttt{CLS} + transformer'' refers to replacing RGCN in \textsc{SIFT} with extra transformer layers.}\label{tab:ablation} 
\end{table}

\subsection{More Discussions}

Besides the specific architecture discussed in $\S$\ref{sec:graph_encode}, our study can be generalized to more knowledge fusion methods. For example, \citet{zhang2020semantics} incorporated semantic role information by combining token embeddings and role type embeddings, which improved the performance of GLUE benchmark over BERT baselines. 
Similar experiments like replacing parsed role sequences with random sequences, are left for future work.   

When it comes to entity knowledge-augmented methods, \citet{Raman2021LearningTD} also observed similar findings as ours that using perturbed KGs can maintain the downstream performance of the original KG though the perturbed KGs are significantly different in terms of semantics and graph structures.
It again demonstrate that the way those methods use knowledge does not align with human priors. 
Both our findings can guide the future work on robust evaluation and explainability analysis of knowledge fusion methods. 

\section{Conclusion}

Our study demonstrates that GLUE tasks can benefit from both trivial graphs and linguistic graphs, indicating that the performance gains of previous fusion methods should not be attributed to linguistic bias entirely.
We argue that comparisons merely between methods with and without knowledge fusion may not be able to capture the whole picture. For example, without baselines considering trivial graph structures, the quality of various fused knowledge may not be accurately assessed. 
More careful evaluations of the effectiveness claims in existing work~\citep[\emph{inter alia}]{sachan-etal-2021-syntax,liu-etal-2020-sentence,peng-etal-2021-structure} may be encouraged in the same spirit.
In addition, tasks and evaluation benchmarks are also crucial to investigate when linguistic structures help.  
Our study contributes to the broader question of how to accurately evaluate models that integrate external knowledge for downstream tasks, such as world or commonsense knowledge~\cite{xu2021does,zhu-etal-2022-knowledge}. 






\section*{Limitations}

We outline two limitations in our study. First, our comparison experiments require high-quality parsers to obtain accurate linguistic graphs. This assumption does not always hold especially for complex semantic parsing. In such case, it is rather difficult to quantify the effect of parsing results on downstream tasks' performance. Note that our constructed trivial graphs need no parser. 
Second, although we systematically evaluate on the GLUE benchmark, more diverse tasks like structure prediction tasks, more knowledge fusion methods, and more linguistic structures like phrase structures~\cite{kong-etal-2015-transforming} remain to be explored in the future work.  



\section*{Acknowledgements}

The authors of this paper were supported by the NSFC Fund~(U20B2053) from the NSFC of China, the RIF~(R6020-19 and R6021-20) and the GRF~(16211520) from RGC of Hong Kong, the MHKJFS~(MHP/001/19) from ITC of Hong Kong and the National Key R\&D Program of China~(2019YFE0198200) with special thanks to HKMAAC and CUSBLT, and the Jiangsu Province Science and Technology Collaboration Fund~(BZ2021065). We also thank the support from the UGC Research Matching Grants~(RMGS20EG01-D, RMGS20CR11, RMGS20CR12, RMGS20EG19, RMGS20EG21).
We would also like to thank Guanlin Li, Xin Liu, Hongming Zhang, Wei Wang, Huan Zhao for their insightful discussions and useful comments.

\bibliography{anthology,custom}
\bibliographystyle{acl_natbib}

\appendix

\section{Statistics of GLUE Benchmark}
\label{sec:app_data_stat}

In Table~\ref{tab:data_statistics}, we list the tasks as well as dataset statistics of the GLUE benchmark. We also show the averaged degrees of parsed semantic graphs for each dataset.  

\begin{table}[h]
\centering
\setlength\tabcolsep{3pt}
\begin{tabular}{llccc}
\hline
Dataset & Task  & \#Train & \#Dev  & Avg |D|\\
\hline
CoLA &  Acceptability & 8.5K & 1K & 1.74\\
MRPC &  Paraphrase  & 2.7K & 409 & 1.91\\
STS-B & Similarity &  5.8K & 1.5k & 1.81\\
SST-2 & Sentiment  &  67K  & 873  & 1.79\\ 
RTE   &  Entailment &  2.5K & 278 & 1.94 \\ 
QNLI &  Entailment  & 105k & 5.5K & 1.91\\
QQP   & Paraphrase &   363K & 40K  & 1.82\\
MNLI &  Entailment  &  392k & 9.8K & 1.89\\ 
\hline
\end{tabular}
\caption{The statistics of evaluation datasets in the GLUE benchmark. \#Train and \#Dev refer to the size of training set and development set respectively.}
\label{tab:data_statistics}
\end{table}

\begin{table}[h]
\centering
\setlength\tabcolsep{3pt}
\begin{tabular}{lcc}
\hline
Dataset & Node Rate  & Edge Rate \\
\hline
CoLA &  93.2\% & 89.0\% \\
MRPC &  81.9\%  & 75.4\%  \\
STS-B & 84.8\% &  79.7\%  \\
SST-2 & 87.8\%  &  80.6\%   \\ 
RTE   &  81.5\% &  74.7\%  \\ 
QNLI &  84.4\%  & 78.6\%  \\
QQP   & 84.2\% &   78.9\% \\
MNLI &  87.2\%  &  80.6\% \\ 
\hline
\end{tabular}
\caption{The ratios of nodes and edges in skeleton graph compared with original parsed graphs.}
\label{tab:skelton_ratio}
\end{table}

\section{Statistics of Skeleton Graphs}
\label{sec:skeleton}

We present the remained ratios of nodes and edges in the skeleton graphs in Table~\ref{tab:skelton_ratio}.




\end{document}